\documentclass[lettersize,journal]{IEEEtran}
\usepackage{amsmath,amsfonts}
\usepackage{algorithmic}
\usepackage{algorithm}
\usepackage{array}
\usepackage[caption=false,font=normalsize,labelfont=sf,textfont=sf]{subfig}
\usepackage{textcomp}
\usepackage{stfloats}
\usepackage{url}
\usepackage{verbatim}
\usepackage{graphicx}
\usepackage{cite}
\usepackage{microtype}
\usepackage{hyperref} 
\usepackage{authblk}
\begin{document}
\title{ROSClaw: A Hierarchical Semantic-Physical Framework for Heterogeneous Multi-Agent Collaboration}
\author{Rongfeng Zhao$^{*}$,  Xuanhao~Zhang$^{*}$, Zhaochen~Guo$^{*}$, Xiang~Shao$^{*}$, Zhongpan Zhu$^{\dagger}$, Bin~He$^{\dagger}$~\IEEEmembership{Senior Member,~IEEE,} and ~Jie ~Chen,~\IEEEmembership{Fellow,~IEEE,}
\thanks{
This work was supported in part by the National Natural Science Foundation of China (No. 2024YFB4709800), in part by the Science and Technology Commission of Shanghai Municipality (No. 2021SHZDZX0100).}
\thanks{Rongfeng Zhao, Xuanhao Zhang, Zhaochen Guo, Xiang Shao are with the Shanghai Research Institute for Intelligent Autonomous Systems, Shanghai 201210, China, and also with the National Key Laboratory of Autonomous Intelligent Unmanned Systems (Tongji University), Shanghai 201210, China, and also with the Frontiers Science Center for Intelligent Autonomous Systems, Shanghai 201210, China. (e-mail: 2311771@tongji.edu.cn; 2510231@tongji.edu.cn; 2411972@tongji.edu.cn; ai@rosclaw.io).}
\thanks{Zhongpan Zhu is with the Shanghai Research Institute for Intelligent Autonomous Systems, Tongji University, Shanghai 201210, China, and also with the National Key Laboratory of Autonomous Intelligent Unmanned Systems (Tongji University), Shanghai 201210, China, and also with the Frontiers Science Center for Intelligent Autonomous Systems, Shanghai 201210, China. (e-mail: 521bergsteiger@tongji.edu.cn).}
\thanks{ Bin He is with the College of Electronics and Information Engineering, Tongji University, Shanghai 201804, China, and alsowith the Shanghai Research Institute for Intelligent Autonomous Systems, Shanghai 201210, China, and also with the National Key Laboratory of Autonomous Intelligent Unmanned Systems (Tongji University), Shanghai 201210, China, and also with the Frontiers Science Center for Intelligent Autonomous Systems, Shanghai 201210, China. (e-mail: hebin@tongji.edu.cn). }
\thanks{ Jie Chen is are with the Shanghai Research Institute for Intelligent Autonomous Systems, Shanghai 201210, China, and also with the National Key Laboratory of Autonomous Intelligent Unmanned Systems (Tongji University), Shanghai 201210, China, and also with the Frontiers Science Center for Intelligent Autonomous Systems, Shanghai 201210, China. (e-mail: chenjie206@tongji.edu.cn). }
\thanks{$^{*}$Equal contribution; $^{\dagger}$Corresponding authors}}




\maketitle

\begin{abstract}
The integration of large language models (LLMs) with embodied agents has improved high-level reasoning capabilities; however, a critical gap remains between semantic understanding and physical execution. While vision–language–action (VLA) and vision–language–navigation (VLN) systems enable robots to perform manipulation and navigation tasks from natural language instructions, they still struggle with long-horizon sequential and temporally structured tasks. Existing frameworks typically adopt modular pipelines for data collection, skill training, and policy deployment, resulting in high costs in experimental validation and policy optimization. To address these limitations, we propose ROSClaw, an agent framework for heterogeneous robots that integrates policy learning and task execution within a unified vision–language model (VLM) controller. The framework leverages e-URDF representations of heterogeneous robots as physical constraints to construct a sim-to-real topological mapping, enabling real-time access to the physical states of both simulated and real-world agents. We further incorporate a data collection and state accumulation mechanism that stores robot states, multimodal observations, and execution trajectories during real-world execution, enabling subsequent iterative policy optimization. During deployment, a unified agent maintains semantic continuity between reasoning and execution, and dynamically assigns task-specific control to different agents, thereby improving robustness in multi-policy execution. By establishing an autonomous closed-loop framework, ROSClaw minimizes the reliance on robot-specific development workflows.  The framework supports hardware-level validation, automated generation of SDK-level control programs, and tool-based execution, enabling rapid cross-platform transfer and continual improvement of robotic skills. Ours project page: \href{https://www.rosclaw.io/}{https://www.rosclaw.io/}.
\end{abstract}


\section{Introduction}
\IEEEPARstart{T}{he} rapid development of LLMs, together with VLA and VLN systems, has enabled robots to perform manipulation and navigation tasks driven by natural language instructions \cite{ref1,ref2,ref3}, improving capabilities in high-level task planning and semantic reasoning. This evolution has further strengthened agents’ abilities in complex decision-making and long-horizon reasoning. With the advancement of embodied intelligence, systems are rapidly transitioning from single-agent task execution \cite{ref4} toward robust heterogeneous multi-agent collaboration \cite{ref5,ref6,ref7,ref8}. In complex and dynamic environments, combining the superior mobility of humanoid robots with the high-precision manipulation capabilities of fixed robotic arms provides highly complementary system-level advantages.

However, extending these capabilities to real-world long-horizon tasks and heterogeneous multi-agent systems reveals a fundamental semantic–physical gap\cite{ref9,ref10}. Existing frameworks typically modularize data collection, policy learning, and deployment into disjoint stages \cite{ref11}. This design introduces semantic and distributional inconsistencies across stages and leads to a strong dependence on manual environment resetting. As a result, multi-policy execution becomes fragile in long-horizon tasks. 

Moreover, many existing hierarchical embodied frameworks \cite{ref12} lack explicit physical constraints, which can lead to incorrect reasoning about system capabilities and the generation of physically infeasible or unsafe task plans \cite{ref13}. Conventional architectures often directly interface high-level semantic APIs with low-level hardware, introducing communication latency and limiting high-frequency closed-loop control. Without runtime physical supervision, small execution errors can quickly accumulate and cascade, ultimately resulting in catastrophic failure in long-horizon tasks. In practice, the absence of a unified abstraction layer forces engineers to perform extensive hardware-specific tuning across heterogeneous platforms, severely limiting scalability and reusability. Furthermore, although large models exhibit strong abstract reasoning capabilities, they lack grounded understanding of physical constraints, particularly the temporal and spatial requirements necessary for reliable action execution.

To address these fundamental limitations, we propose ROSClaw, a hierarchical semantic–physical agent framework specifically designed for heterogeneous multi-agent collaboration in dynamic environments. Unlike conventional systems that rely on manual intervention or static planners, ROSClaw adopts a unified three-layer architecture spanning the information space, software system, and physical world, rather than treating tasks as isolated execution pipelines. Through an asynchronous decoupling mechanism, the framework separates low-frequency semantic planning from high-frequency physical control, enabling stable interaction across different temporal scales. To abstract away the heterogeneity of underlying hardware, ROSClaw introduces an Online Tool Pool that automatically maps abstract instructions into executable software calls, enabling cross-platform generalization under a “train once, deploy everywhere” paradigm. Prior to execution, the framework leverages a digital twin engine together with e-URDF-based physical constraints to perform forward dynamics simulation, filtering infeasible commands generated by large models prior to execution.

From a system-level perspective, this work advances a new paradigm that models embodied intelligence as a resource-driven closed-loop process centered on shared tools and reusable skills. We design a unified pipeline that integrates e-URDF-based physical constraints with data collection and state accumulation mechanisms. This architecture transforms joint states, multimodal observations, and task execution experiences into reusable skills and structured datasets. By feeding such physically grounded data back into the cognitive layer, the framework enables self-organizing learning and improves the alignment between semantic reasoning and physical execution. The resulting data feedback mechanism converts physical interactions into continuously accumulated knowledge, reducing reliance on manual annotation and demonstration. In practice, this mechanism can significantly reduce annotation and demonstration costs from days to hours, enabling more efficient system adaptation and supporting the transition from isolated task execution toward continual self-improvement.

The main contributions of this work are summarized as follows:
\begin{enumerate}
\item{ROSClaw system architecture. We propose a semantic–physical framework spanning the information space, software system, and physical world, which decouples the macro-level knowledge engine of large models from low-level high-frequency physical control. The framework enables unified abstraction and coordination across multiple agents, and integrates policy learning with long-horizon task execution within a single agent loop, ensuring contextual semantic consistency from planning to execution.}
\item{e-URDF-based physical constraints with data collection and state accumulation. We introduce e-URDF-based physical constraints together with a data collection and state accumulation mechanism. A headless digital twin sandbox is constructed based on Isaac Lab, in which collision detection and joint torque validation are performed using e-URDF representations of heterogeneous agents prior to dispatching commands to the physical resource pool, thereby enforcing strict physical boundaries. This mechanism continuously records execution trajectories and task states of agents and stores them in a local resource pool for subsequent reuse and policy optimization.}
\item{Real-world validation on heterogeneous multi-agent systems. We validate the proposed framework on a distributed hardware platform consisting of a humanoid robot, a fixed arm manipulator, and a mobile manipulation system. The system demonstrates robust collaboration in shared environments, including mobile manipulation by the mobile robotic arm platform, navigation by the humanoid robot, and precise operations by the fixed robotic arm manipulator. Compared to conventional workflows that rely on manual hardware testing and iterative programming, ROSClaw enables tool-driven autonomous hardware validation and SDK-level program synthesis, thereby reducing human intervention.}
\end{enumerate}
\section{Related Work}
 {\bf{LLM-based multi-agent systems (MAS).}} Early multi-agent systems based on LLMs primarily focused on software and digital environments, adopting centralized or hierarchical coordination structures. For instance, frameworks such as MetaGPT \cite{ref6} and CAMEL construct top-down task allocation pipelines through predefined procedures and role assignments. Similarly, the cross-environment evaluation framework Crab \cite{ref14} leverages multimodal models to generate centralized instructions for GUI automation, demonstrating strong performance on highly structured tasks.  

 To improve adaptability in dynamic environments, recent work explores decentralized coordination through peer-to-peer communication. Approaches such as multi-agent debate and feedback mechanisms \cite{ref15,ref16} reduce hallucination in individual models and improve collective decision-making by enabling iterative consensus formation. 

 More recently, these coordination paradigms have been extended to embodied settings. Dual-agent systems based on point-to-point communication have been applied to search and navigation tasks in multi-room environments \cite{ref17}. The RoCo framework \cite{ref18} introduces a decentralized debate-based structure for multi-arm collaboration, leveraging LLM-based spatial reasoning to assist low-level trajectory planning. In addition, some studies incorporate decentralized coordination nodes to control heterogeneous robotic systems, including aerial vehicles, legged robots, and wheeled manipulators, for navigation and manipulation tasks \cite{ref19,ref20}. 

 Despite these advances, most existing embodied multi-agent systems remain constrained by predefined communication structures. Moreover, inter-agent collaboration is typically limited to high-level task planning via API-level interactions, lacking tight integration with low-level physical control in real-world environments.

 {\bf{Embodied intelligence and multi-robot collaboration.}} Traditional heterogeneous multi-robot systems (MRS) typically rely on manually designed task decomposition frameworks, heuristic resource allocation strategies, or specifically trained reinforcement learning and communication protocols. These approaches often struggle to generalize in dynamic environments and across heterogeneous robotic platforms (e.g., cross-modal coordination among aerial, legged, and wheeled robots).

 Recent advances in LLMs have enabled a new paradigm that leverages commonsense reasoning for zero-shot task planning. Early methods primarily adopt open-loop planning, generating full action sequences in advance \cite{ref21}, while subsequent approaches incorporate feedback to enable closed-loop planning \cite{ref22,ref23}. Methods such as Code as Policies \cite{ref24} and VoxPoser \cite{ref25} improve high-level task decomposition, whereas hierarchical frameworks—including Hi Robot \cite{ref26}, HAMSTER \cite{ref27}, and Coa-VLA \cite{ref28}—introduce structured subtask abstraction and plan–verify mechanisms for adaptive coordination.

 To bridge the gap between semantic planning and physical execution, VLA models such as PaLM-E \cite{ref29} and OpenVLA \cite{ref30} provide end-to-end alignment between perception, language, and action. Additional methods, including Reflect-VLM \cite{ref31} and SERP \cite{ref32}, improve robustness through replanning mechanisms. Some works also integrate reinforcement learning into agent systems \cite{ref33}, leveraging feedback from real interactions such as user dialogues and tool invocations, where subsequent states are used as supervision signals. This enables agents to learn during deployment through unimodal data collection and state accumulation. In addition, Li et al. \cite{ref34} propose Entangled Action Pairs (EAP), which couple forward actions with recovery behaviors to support self-resetting data collection.

 Despite these advances, existing embodied frameworks remain limited in real-world execution. End-to-end VLA models are prone to error accumulation in long-horizon collaborative tasks. Moreover, both zero-shot planners and hierarchical verification frameworks typically perform supervision only at discrete subtask boundaries, lacking continuous process-level feedback during low-level execution. In heterogeneous multi-robot settings, large models are often used as centralized schedulers, while execution is delegated to fragmented pre-programmed APIs, resulting in weak coupling between high-level reasoning and physical control.
\begin{figure*}[h]
\centering
\includegraphics[width=0.75\textwidth]{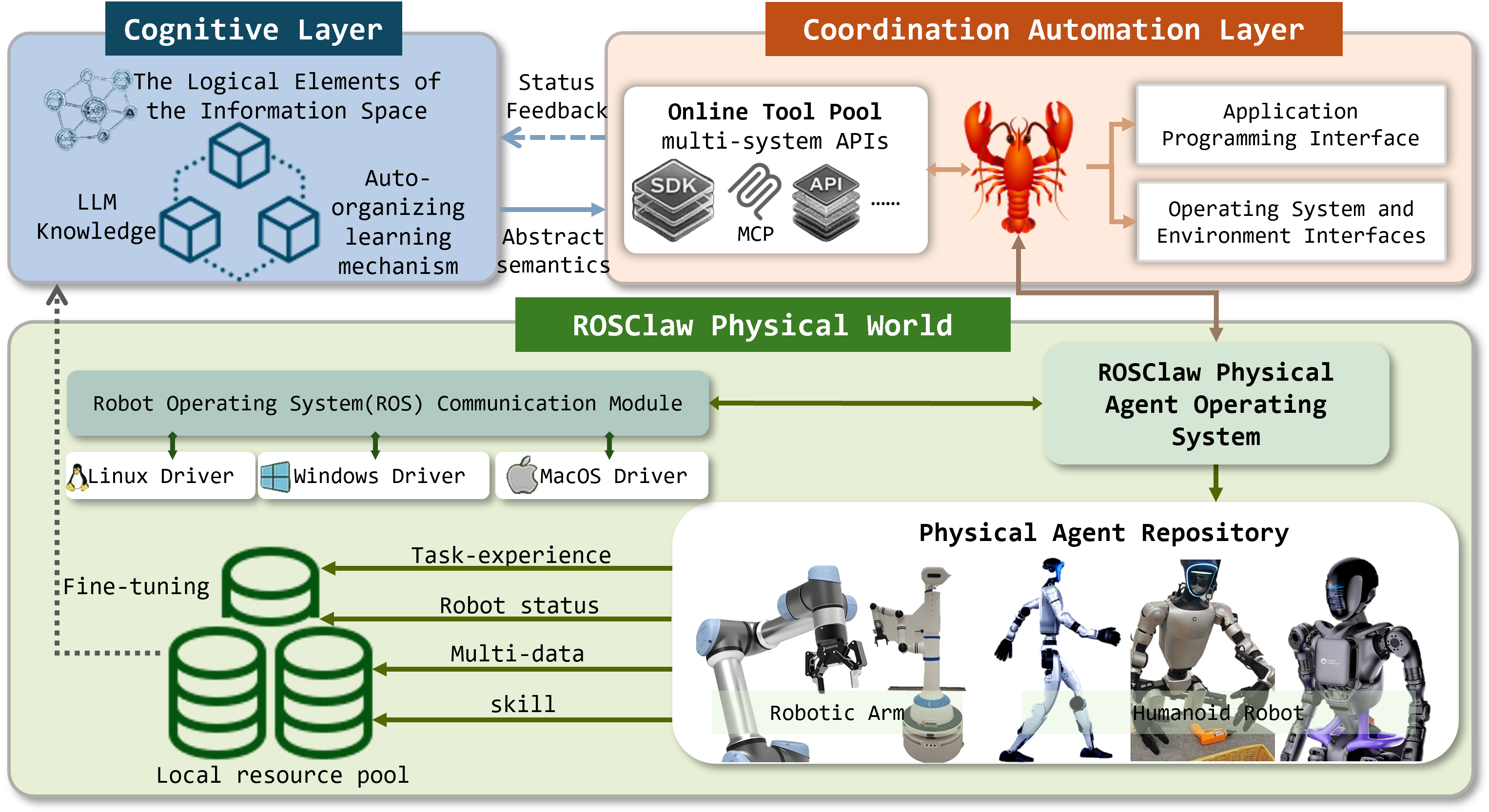}
\caption{The ROSClaw framework adopts a three-layer semantic–physical architecture to bridge high-level cognitive reasoning and low-level physical control. The cognitive layer relies on LLM knowledge graphs and logical elements in the digital space to support macro-level task understanding. The coordination automation layer abstracts hardware heterogeneity through an Online Tool Pool and enables task environment activation. The ROSClaw physical world provides unified control over heterogeneous robotic agents while continuously accumulating multimodal observations, robot states, and reusable skills within a Local Resource Pool. The accumulated interaction experience is fed back to the cognitive layer, forming a closed-loop process that supports continual system evolution and cross-task knowledge reuse.}\label{fig1}
\end{figure*}
\section{Framework Overview}
Modern embodied intelligence systems face a fundamental mismatch between high-level cognitive reasoning and low-level high-frequency physical control during real-world deployment. This mismatch arises from two key factors. First, large models primarily operate in the semantic and linguistic domain and lack explicit representations of physical dynamics, control constraints, and temporal dependencies required for reliable execution. Second, heterogeneous robotic platforms, characterized by fragmented SDKs and non-standardized interfaces, hinder unified integration and efficient coordination across agents. More critically, existing frameworks are typically designed as static, open-loop pipelines that fail to convert physical interactions into reusable knowledge, limiting the ability of embodied systems to continuously adapt and improve in real-world environments.

 To address these challenges, we propose ROSClaw, an operating system–level framework for heterogeneous multi-agent embodied systems. As illustrated in Fig.\ref{fig1}, ROSClaw establishes a three-layer semantic–physical architecture spanning the information space, software system, and physical world. Rather than treating tasks as isolated execution pipelines, the framework models embodied intelligence as a resource-driven, closed-loop process in which perception, reasoning, and execution are tightly coupled with data accumulation and continual learning. Two core abstractions are introduced to support this process, including the Online Tool Pool, which bridges semantic reasoning and executable software actions, and the Local Resource Pool, which accumulates physical interaction data and reusable skills. Together, these components form a unified pipeline that enables continuous system evolution through structured resource flow.
\subsection{Cognitive Layer}
The cognitive layer consists of large language and vision–language models that function as the high-level decision engine. This layer is responsible for low-frequency task decomposition, long-horizon reasoning, and semantic understanding of the environment. Unlike conventional one-way instruction pipelines, the cognitive layer operates under a bidirectional interaction paradigm. It generates structured semantic representations and task plans that are passed to the coordination layer, while simultaneously receiving execution feedback, environmental states, and interaction outcomes from the physical world.

This bidirectional feedback loop enables the gradual alignment of semantic reasoning with physical execution, thereby mitigating inconsistencies introduced by purely text-based pretraining. Through iterative interaction with real-world data, the system develops grounded spatiotemporal understanding of the environment rather than relying solely on abstract symbolic reasoning. As a result, the cognitive layer evolves from a static planner into an adaptive reasoning module that continuously refines its decisions based on embodied experience.
\subsection{The coordination automation layer}
The coordination layer is centered on the OpenClaw system and is primarily responsible for the abstraction of task logic and the orchestration of automated tools. To enable seamless interoperability across heterogeneous systems, this layer constructs an Online Tool Pool. This pool aggregates various robot Software Development Kits (SDKs), Model Context Protocols (MCPs), and multi-system API interfaces. It functions as a “digital dictionary” bridging the cognitive layer and the execution layer, allowing abstract instructions generated by the cognitive layer to be automatically translated and mapped into concrete software invocation procedures. Furthermore, the coordination layer incorporates a strict e-URDF-based physical interception and scheduling mechanism. As illustrated in Figure 2, before actions are dispatched to the physical world, forward dynamics simulation and collision detection are conducted through a digital twin engine, thereby ensuring physical feasibility.
\begin{figure*}[h]
\centering
\includegraphics[width=0.9\textwidth]{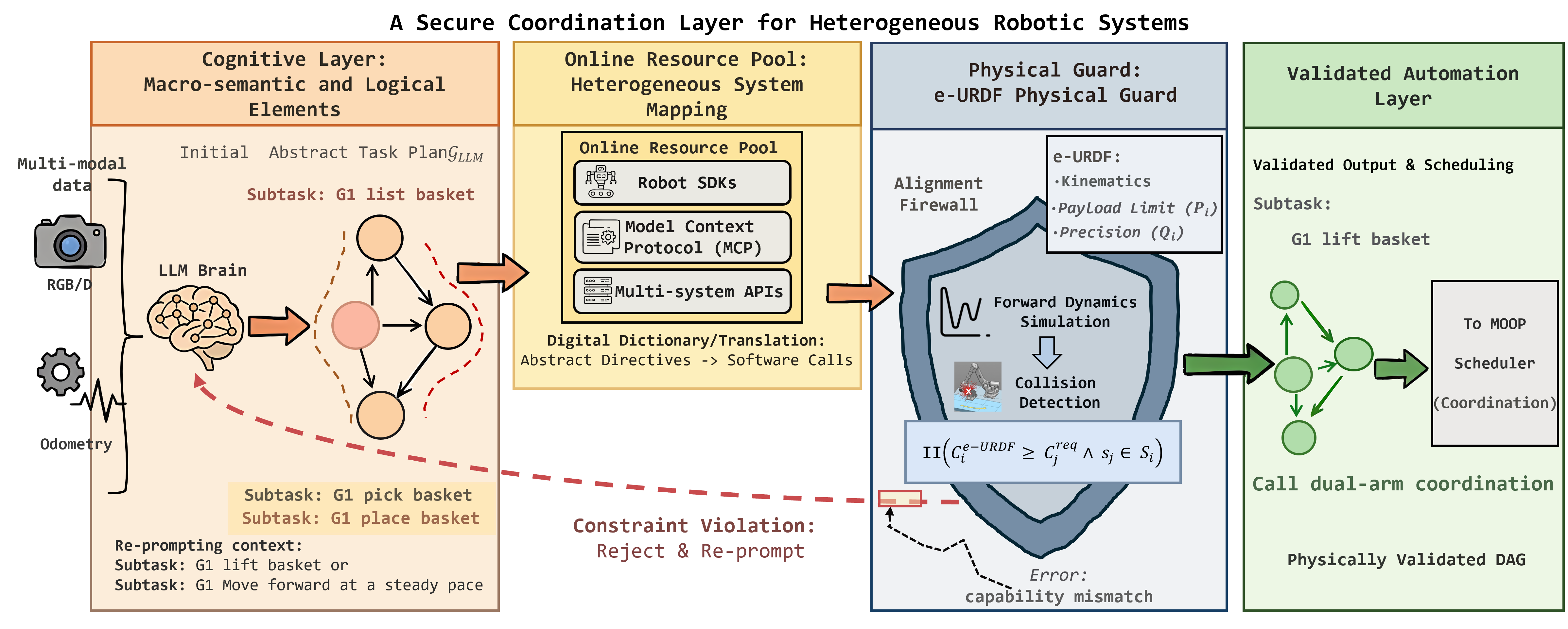}
\caption{e-URDF-based physical firewall. Heterogeneous system resources (SDKs, MCPs, and APIs) are aggregated into an Online Tool Pool to enable the transformation of abstract instructions into executable operations. A strict e-URDF-based physical safeguarding mechanism for heterogeneous agents is adopted, leveraging forward dynamics simulation and collision detection in Isaac Lab to ensure physical feasibility prior to scheduling.}\label{fig2}
\end{figure*}

\subsection{ROSClaw physical world}
The ROSClaw physical world interfaces with high-frequency robot operating systems (ROS) and platform-specific drivers to provide unified control over heterogeneous agents. This abstraction enables policies generated at higher levels to generalize across platforms under a unified interface.

One of the core innovations of this architecture lies in the construction of a local resource pool. During execution, ROSClaw is capable of invoking tool interfaces from the online resource pool to connect with physical-world sensors and acquire data, while simultaneously performing real-time, high-frequency collection of the robot’s physical states, thereby enabling infant-inspired autonomous learning and prediction. The resulting multimodal observations, together with successful long-horizon execution trajectories, are standardized and consolidated into the Local Resource Pool, forming reusable skill representations that support continual policy refinement and cross-scenario generalization.

The accumulated physical experience, scene data, and high-quality tokens stored locally not only support self-iteration of the local agent but can also be uploaded to a cloud-based model resource pool for large-scale model fine-tuning or training. This mechanism of data collection and state accumulation transforms physical interactions into continuously growing knowledge assets, substantially reducing the cost of manual annotation and establishing a foundational framework for cross-scenario intelligent emergence in future embodied AI systems.
\section{ROSClaw Framework Validation}
We evaluate the ROSClaw framework in a complex real-world environment involving four heterogeneous robots performing multi-step tasks. These robots are built on platforms from different manufacturers, employ different drivers and development SDKs, and collaboratively operate within the same environment to accomplish tasks specified directly through natural language descriptions.

\subsection{Temporal Task Multi-Robot Collaborative Operation Validation}
\begin{figure}[h]
\centering
\includegraphics[width=0.5\textwidth]{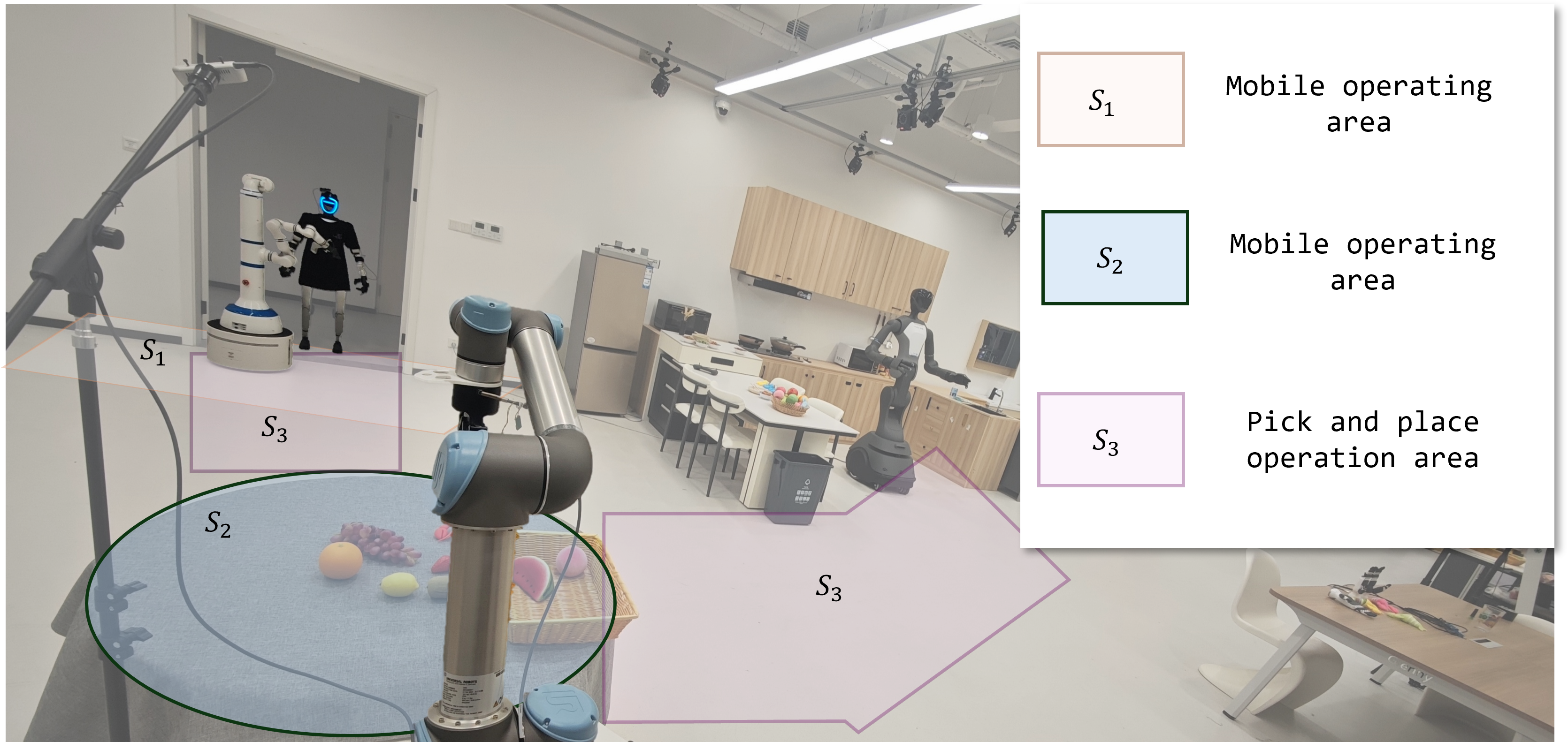}
\caption{Real-World Environment for Collaborative Tasks. In the physical environment, $S_1$denotes the mobile manipulation region, $S_2$represents the localized grasping region, and $S_3$ corresponds to the mobile navigation region. Each embodied agent in the physical world is constrained to operate only within a subset of these regions.}\label{fig3}
\end{figure}

Scenario: The evaluation is conducted in a smart home environment composed of a kitchen and a living room, with an approximate area of 60 square meters. The scene includes three tables, one sink, six cabinets, and one refrigerator, as illustrated in Fig.\ref{fig3}.
The proposed ROSClaw framework provides a unified solution for heterogeneous multi-agent collaboration and advances embodied intelligence toward a closed-loop learning paradigm. By establishing a semantic–physical architecture spanning the information space, software system, and physical world, the framework bridges the gap between high-level reasoning and low-level control, enabling robust execution of long-horizon tasks in real-world environments. From a system perspective, ROSClaw models embodied intelligence as a resource-driven closed-loop process. The integration of e-URDF-based physical constraints, digital twin validation, and multi-objective scheduling ensures safe and feasible execution, while the Task Execution Supervision (TES) mechanism and data feedback loop support continual policy refinement. Building upon this foundation, the Local Resource Pool accumulates and organizes execution data, skills, and interaction experiences for reuse.. his mechanism improves system adaptability and provides a pathway for incorporating physically grounded data into large model training, supporting generalization across tasks and environments.

Task: Three robots participate in this collaborative task and are divided into two groups. The mobile manipulation group consists of the mobile robotic arm and the humanoid robot, while the harvesting and navigation group consists of the fixed robotic arm and humanoid robot. Due to workspace constraints, each robot can only access a subset of regions, while some robots are capable of executing tasks across multiple regions and support different functionalities. Specifically, the mobile robotic arm is equipped with a wheeled base and a two-finger gripper, enabling mobile manipulation, whereas is equipped with a head-mounted camera that allows it to perceive fruits on the living room table and transport them to the kitchen sink through navigation. Detailed configurations of the robots and environment are provided in the supplementary materials (\href{https://www.rosclaw.io/demo}{video}) .

\begin{figure*}[h]
\centering
\includegraphics[width=0.85\textwidth]{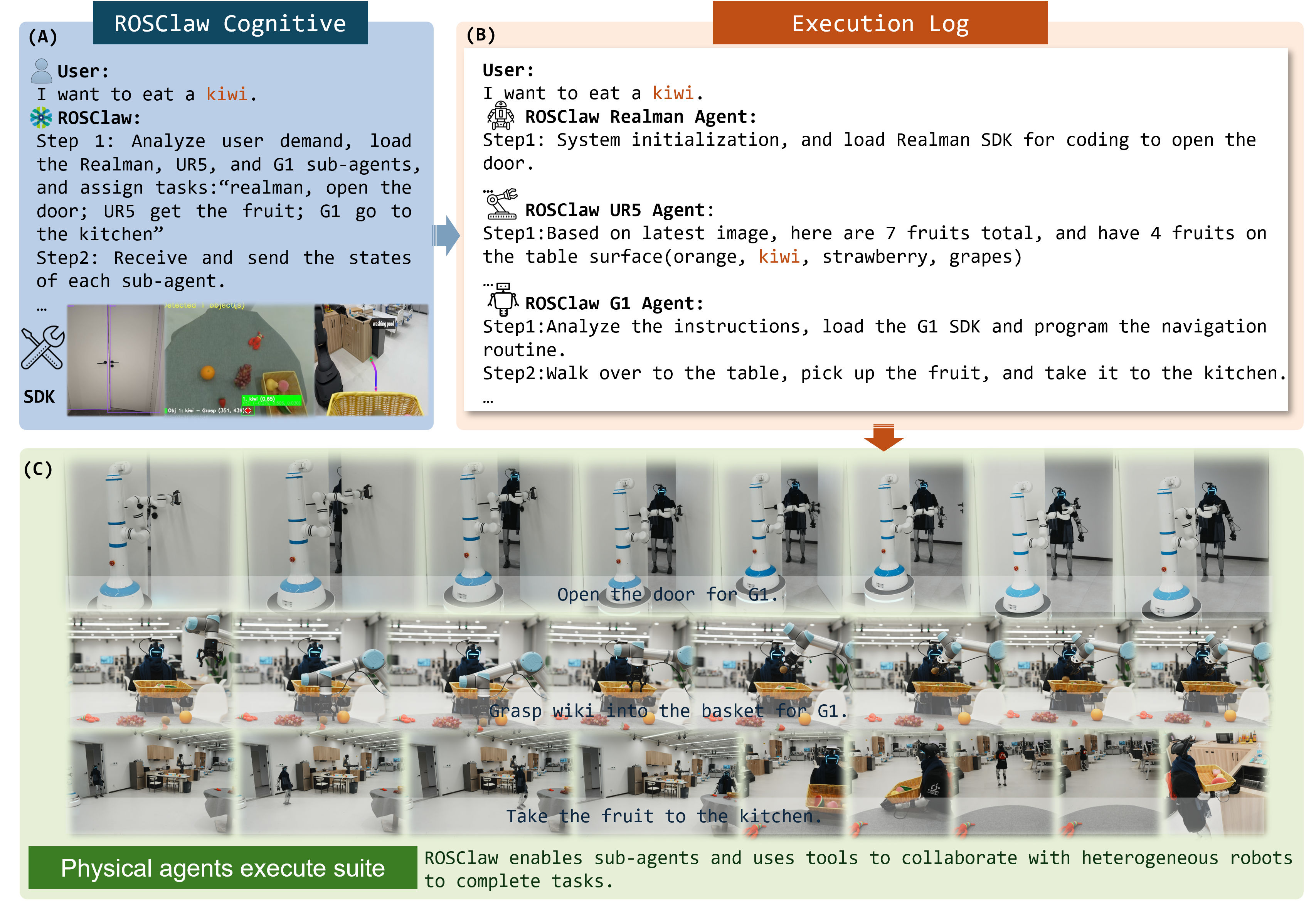}
\caption{Heterogeneous Multi-Agent Collaboration. In (A), ROSClaw receives user requirements, initializes the sub-agents, assigns tasks to each agent, and simultaneously exchanges state information with them. In (B), the log outputs generated during task execution by each sub-agent are presented. In (C), the physical-world execution is illustrated: the mobile robotic arm approaches the doorway and opens the door; the humanoid robot enters and moves toward the harvesting area; the fixed robotic arm transfers the fruit; and the humanoid robot carries the fruit basket to the kitchen.}\label{fig4}
\end{figure*}

First, ROSClaw monitors and analyzes user requirements, followed by an analysis of each robot’s accessible regions to assign tasks accordingly. During task execution, in the mobile manipulation group, the mobile robotic arm approaches the doorway, uses its robotic arm to open the door, and then relocates. Subsequently, humanoid robot enters the living room, approaches the table, picks up the fruit basket, and waits for the fixed robotic arm to place fruits into the basket. Meanwhile, ROSClaw transmits environment perception results from a VLM to the arm agent, including the quantity and categories of fruits on the living room table, as shown in Fig.\ref{fig4}(A). It then invokes the DINO-X API from the online resource pool to send appropriate grasping candidate points to the fixed robotic arm. Upon receiving the perception and grasping information from ROSClaw, the fixed robotic arm in the harvesting and navigation group uses its two-finger gripper to pick up the user-specified kiwi and place it into the basket, completing the inter-group collaborative interaction. Finally, the humanoid robot follows the navigation trajectory from the living room table to the kitchen sink, provided by ROSClaw, and transports the fruit basket to the sink. The execution of this collaborative task by heterogeneous robots in the physical world is illustrated in Fig.\ref{fig4}(C).
\subsection{Validation of e-URDF and the Data Collection and State Accumulation Mechanism}
To evaluate the effectiveness of the e-URDF-based physical safeguarding mechanism and the data collection and state accumulation mechanism within the ROSClaw framework, we design two task scenarios involving two groups: the fixed robotic arm operation group and a gimbal-based multi-agent orchestration group.

\begin{figure*}[h]
\centering
\includegraphics[width=0.9\textwidth]{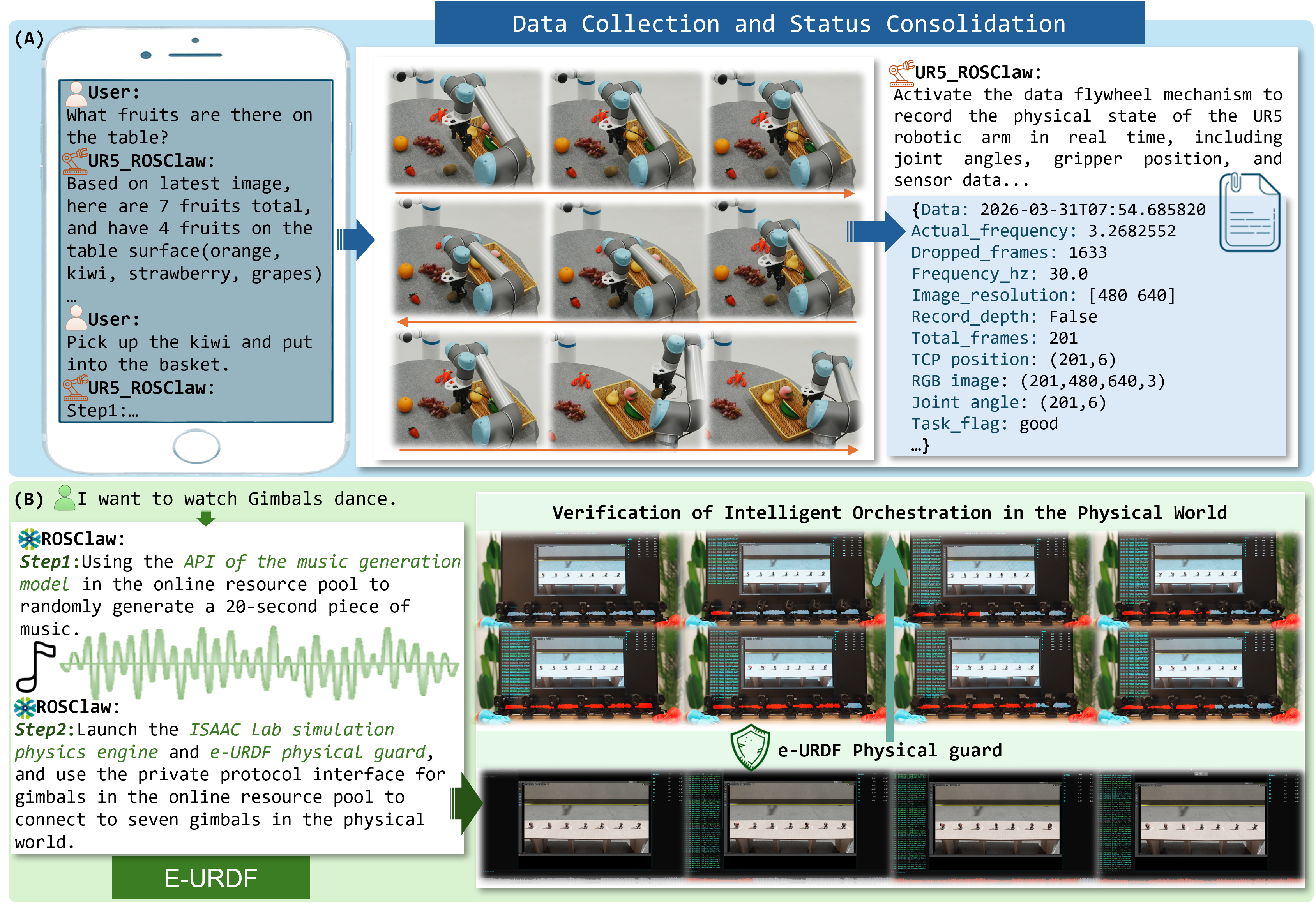}
\caption{Validation of e-URDF-based physical safeguarding and the data collection and state accumulation mechanism. In (A), a mobile user interacts with ROSClaw to activate perception and manipulation by the arm agent in the physical world, while simultaneously triggering the data collection and state accumulation mechanism to record agent states and environmental perception data. In (B), ROSClaw generates music based on user instructions, initiates e-URDF-based physical safeguarding to orchestrate a dance routine, and drives the physical system to enable coordinated dancing of seven gimbal units.}\label{fig5}
\end{figure*}

First, the user activates the ROSClaw, the arm agent on the client side and engages in interactive dialogue. After the user queries “What fruits are there on the table?”, ROSClaw activates the RealSense camera, while the agent leverages the VLM API from the online resource pool to perform tabletop environment perception. The system responds to the user in a structured list format, including the color, category, and 3D positions of the fruits relative to the fixed robotic arm, and subsequently engages the user for the next-step requirement. Then, based on the user’s updated request, the system decomposes the task into executable steps and transmits the task execution logs back to the user. Meanwhile, key execution trajectories and perception results are also reported, enabling the user to remotely monitor the agent’s state and task progress in the physical world, as illustrated in Fig.\ref{fig5}(A).
We validate the effectiveness of e-URDF-based physical safeguarding through a designed multi-agent orchestration task. After the user queries “I want to watch Gimbals dance.”, ROSClaw instantiates a Gimbal agent and invokes a music generation model via the online resource pool to produce approximately 20 seconds of music. It then launches Isaac Lab and creates an MCP service that interfaces with the physical gimbal units, which is subsequently registered in the online resource pool.

Next, the agent loads the URDF models of the gimbals in Isaac Lab and continuously verifies the physical safety of the automatically generated choreography through simulation. Meanwhile, leveraging the gimbal interface from the online resource pool, ROSClaw connects to seven physical gimbal units and executes the orchestrated dance motions in the real world, as illustrated in Figure 5(B). Extensive experimental results demonstrate that, compared to traditional multi-agent orchestration development pipelines, ROSClaw reduces the time required to generate coordinated multi-gimbal dance behaviors to approximately three minutes, with human involvement limited to the initial instruction specification.
\section{Conclusion}
This paper presents ROSClaw, a hierarchical semantic–physical framework for heterogeneous robotic systems that bridges the gap between high-level reasoning in large language models and low-level physical control in real-world environments. By introducing a unified three-layer architecture spanning the information space, software system, and physical world, ROSClaw addresses the fragmentation of task execution commonly observed in existing embodied intelligence frameworks.

To enable unified scheduling across heterogeneous platforms, we introduce an Online Tool Pool that provides standardized interfaces for mapping abstract semantic instructions to executable software calls. To ensure safety and physical feasibility, we further propose an e-URDF-based physical safeguard mechanism, which validates candidate actions through forward dynamics simulation and collision checking in a digital twin environment prior to execution. In addition, the physical execution layer incorporates a data collection and state accumulation mechanism that continuously records robot states, multimodal observations, and long-horizon execution trajectories, and consolidates them into a Local Resource Pool for reuse. Experiments in real-world environments demonstrate that ROSClaw supports effective coordination across heterogeneous agents, including humanoid robots and multi-degree-of-freedom manipulators, under cross-modal and cross-region constraints.

Despite these advantages, several limitations remain. First, current evaluations are primarily conducted in structured or semi-dynamic environments, and a systematic framework for handling complex uncertainties—such as high-frequency disturbances, perception noise, and model stochasticity—has yet to be established. Second, although the Local Resource Pool enables data accumulation, it does not yet form a closed-loop learning system that tightly integrates data acquisition with autonomous policy optimization.

Future work will focus on enhancing robustness and learning capability in real-world deployment. Specifically, we will investigate probabilistic modeling and dynamic obstacle avoidance to improve performance in highly unstructured environments. In addition, we aim to leverage the accumulated multimodal interaction data to enable continual learning during deployment, establishing a closed-loop pipeline in which physical experience directly contributes to the adaptation of high-level cognitive models. This direction provides a pathway toward scalable and continuously improving embodied intelligence systems.


 
%
\bibliographystyle{IEEEtran}

\begingroup
\sloppy
\bibliography{myref}
\endgroup
\newpage
\vspace{11pt}
\vfill

\end{document}